\documentclass[11pt,a4paper]{article}
\usepackage[hyperref]{acl2017}
\usepackage{times}
\usepackage{latexsym}
\newcommand{\mCBP}{CB online~}
\newcommand{\mRANLP}{\emph{anyG~}}
\newcommand{\mRANLPsingle}{singletonG~}
\newcommand{\mA}{{\em singleton}~}
\newcommand{\mB}{{\em any}~}
\newcommand{\mC}{{\em singleton+any}~}
\newcommand{\mD}{{\em multi}~}
\newcommand{\mE}{{\em multi+any}~}
\newcommand{\mCBfeats}{singleton CB~}

\usepackage{multicol, listings}
\usepackage{booktabs}
\usepackage{mathabx}
\usepackage[colorinlistoftodos]{todonotes}

\usepackage{multirow}
\newcommand{\blue}{\textcolor{blue}}

\usepackage{url}

\aclfinalcopy

\title{It Takes Nine to Smell a Rat: \\
Neural Multi-Task Learning for Check-Worthiness Prediction}

\author{Slavena Vasileva \\
  Sofia University \\
  {\tt slav.vasileva@gmail.com} \\\And
    Pepa Atanasova \\ 
    University of Copenhagen\\
    {\tt pepa@di.ku.dk} \\\And
    Llu\'{i}s M\`{a}rquez \\ 
    Amazon Core ML\\
    {\tt lluismv@amazon.com} \\\AND
    Alberto Barr\'{o}n-Cede\~{n}o \\ 
    Universit\`{a} di Bologna\\
    {\tt a.barron@unibo.it} \\\And
    Preslav Nakov \\ 
    Qatar Computing Research Institute, HBKU\\
    {\tt pnakov@hbku.edu.qa} \\
}

\date{}

\begin{document}
\maketitle
\begin{abstract}
  We propose a multi-task deep-learning approach for estimating the check-worthiness of claims in political debates. Given a political debate, such as the 2016 US Presidential and Vice-Presidential ones, the task is to predict which statements in the debate should be prioritized for fact-checking. While different fact-checking organizations would naturally make different choices when analyzing the same debate, we show that it pays to learn from multiple sources simultaneously (PolitiFact, FactCheck, ABC, CNN, NPR, NYT, Chicago Tribune, The Guardian, and Washington Post) in a multi-task learning setup, even when a particular source is chosen as a target to imitate. Our evaluation shows state-of-the-art results on a standard dataset for the task of check-worthiness prediction.
\end{abstract}

\section{Introduction}

Recent years have seen the explosion of fake news, rumors, false claims, distorted facts, half-true statements, and propaganda, which are spreading primarily in social media, but also via standard news broadcasters.
This trend became particularly evident during the 2016 US Presidential campaign, which was the turning point that attracted wide public attention to the problem. 
By then, a number of organizations, e.g., FactCheck\footnote{\url{http://www.factcheck.org/}} and Snopes\footnote{\url{http://www.snopes.com/}} among many others, launched fact-checking initiatives. 
Yet, this proved to be a very demanding manual effort, and only a relatively small number of claims could be fact-checked. Thus, it is important to prioritize what to check.

\noindent The task of detecting check-worthy claims has been recognized as an important stage in the process of fully automatic fact-checking. According to \newcite{vlachos2014fact} this is a multi-step process that
(\emph{i})~extracts statements to be fact-checked,
(\emph{ii})~constructs appropriate questions,
(\emph{iii})~obtains the answers from relevant sources, and
(\emph{iv})~reaches a verdict using these answers. \newcite{Hassan_thequest:2015} presented a similar vision, and in a follow up work they made check-worthiness an integral part of an end-to-end fact-checking system \newcite{Hassan:2017:CFE:3137765.3137815}.

Here, we approach the problem of mimicking the selection strategy of several renowned fact-checking organizations such as PolitiFact, FactCheck, ABC, CNN, NPR, NYT, Chicago Tribune, The Guardian, and The Washington Post. An important characteristic of this setup is that, perhaps due to editorial policies, fact-checking organizations often select different claims for the same text, with little overlap in their choices (see Tables~\ref{table:agreement} and \ref{table:exampleA}). 
Yet, it has been previously shown that it might be beneficial to learn from the union of the selections by multiple fact-checking organizations \cite{gencheva-EtAl:2017:RANLP}. 
Thus, we propose a multi-task deep learning framework, in which we try to predict the choice of each and every fact-checking organization simultaneously.
We show that, even when the goal is to mimic the choice of one particular fact-checking organization, it is beneficial to leverage on the choices by multiple such organizations.
The evaluation results on a standard dataset show state-of-the-art results.

The remainder of this paper is organized as follows. Section~\ref{sec:related} provides an overview of related work. Section~\ref{sub:data} describes the used dataset. Section~\ref{sec:method} describes our method and features. Section~\ref{sec:experiments} presents the experiments and the evaluation results. Finally, Section~\ref{sec:conclusion} concludes and points to some possible directions for future work.

\section{Related Work}
\label{sec:related}

The proliferation of false information has attracted a lot of research interest recently. This includes challenging the truthiness of news~\cite{brill2001online,Hardalov2016,Potthast2018}, of news sources~\cite{D18-1389,source:multitask:NAACL:2019}, 
and of social media posts~\cite{Canini:2011,Castillo:2011:ICT:1963405.1963500,PlosONE:2016}, 
as well as studying credibility, influence, bias, and propaganda~\cite{Ba:2016:VERA,Chen:2013:BIW:2492517.2492637,Mihaylov2015FindingOM,Kulkarni:2018:EMNLP,D18-1389,InternetResearchJournal:2018,Barron:19,EMNLP2019:propaganda:finegrained,EMNLP2019:tanbih}. 

Research was facilitated by shared tasks such as the SemEval 2017 and 2019 tasks on Rumor Detection \cite{derczynski-EtAl:2017:SemEval,gorrell-etal-2019-semeval}, the CLEF 2018 and 2019 CheckThat! labs \cite{clef2018checkthat:overall,clef-checkthat:2019,CheckThat:ECIR2019}, which featured tasks on automatic identification \cite{clef2018checkthat:task1,clef-checkthat-T1:2019} and verification \cite{clef2018checkthat:task2,clef-checkthat-T2:2019} of claims in political debates, the FEVER 2018 and 2019 task on Fact Extraction and VERification~\cite{thorne-EtAl:2018:N18-1}, and the SemEval 2019 task on Fact-Checking in Community Question Answering Forums~\cite{mihaylova-etal-2019-semeval}, among others.

The interested reader can learn more about ``fake news'' from the overview by \newcite{Shu:2017:FND:3137597.3137600}, which adopted a data mining perspective and focused on social media.
Another recent survey  \cite{thorne-vlachos:2018:C18-1} took a fact-checking perspective on ``fake news'' and related problems.
Yet another survey was performed by~\newcite{Li:2016:STD:2897350.2897352}, and it covered truth discovery in general.
Moreover, there were two recent articles in \emph{Science}:
\newcite{Lazer1094} offered a general overview and discussion on the science of ``fake news'', while
\newcite{Vosoughi1146} focused on the  proliferation of true and false news online.

The first work to target check-worthiness estimation, i.e.,~predicting which sentences in a given input text should be prioritized for fact-checking, was the ClaimBuster system~\cite{Hassan:15}. It is trained on data that was manually annotated by students, professors, and journalists, where each sentence was marked as \textit{non-factual}, \textit{unimportant factual}, or \textit{check-worthy factual}. 
The system used an SVM classifier and features such as sentiment, TF.IDF representations, part-of-speech tags, and named entities. 

\noindent In our previous work \cite{gencheva-EtAl:2017:RANLP}, we used debates from the 2016 US Presidential Campaign and fact-checking reports by professional journalists; we use this same dataset here.
Beside most of the features borrowed from ClaimBuster, our model paid special attention to the context of each sentence. This includes whether it is part of a long intervention by one of the debate participants and its position within such an intervention. We predicted both (\emph{i})~whether any of the fact-checking organizations would select the target sentence, and also (\emph{ii})~whether a specific fact-checking organization would select it.
There was also a lab on fact-checking at CLEF 2018 and 2019 \cite{clef2018checkthat:task1,clef-checkthat-T1:2019}, which was partially based on a variant of this data, but it focused on one fact-checking organization, unlike our multi-source setup here.

\newcite{Patwari:17} also focused on the 2016 US Election campaign. Their setup asks to predict whether any of the fact-checking organizations would select the target sentence. They used a boosting-like model that takes SVMs focusing on different clusters of the dataset and the final outcome is considered as that coming from the most confident classifier. The features considered range from LDA topic-modeling to part-of-speech (POS) tuples and bag-of-words representations.

Other claim monitoring tools include FactWatcher \cite{Hassan2014DataIF} and DisputeFinder~\cite{ennals2010highlighting}. FactWatcher classifies claims as situational facts, one-of-the-few, or prominent streaks. It checks whether a new text triggers some of the three types of claims, treating the sentences in the text as sequential data. DisputeFinder mines the Web for already-verified claims. Both maintain a growing database of facts and known claims.

Beyond the document context, it has been proposed to mine check-worthy claims on the Web. For example, \newcite{ennals2010disputed} searched for linguistic cues of disagreement between the author of a statement and what is believed, e.g., ``\emph{falsely claimed that X}''. 
The claims matching the patterns would then go through a classifier. This procedure can be used to acquire a dataset of disputed claims.

Given a set of disputed claims, \newcite{ennals2010highlighting} looked for new claims on the Web that entail the ones that have already been collected. Thus, the task can be reduced to recognizing textual entailment~\cite{dagan2009recognizing}.

\noindent \newcite{de2008finding} also looked for contradictions in text. They tried to classify the contradictions that can be found in a piece of text in two categories ---those occurring via antonymy, negation, and date/number mismatch, and those arising from different world knowledge and lexical contrasts. The features that are selected for the task of contradiction detection include polarity, numbers, dates and time, antonymy, factivity, modality, structural, and relational features. 

Finally, \newcite{le2016towards} used deep learning. They 
argued that the top terms in claim vs. non-claim sentences are highly overlapping in content, 
which is a problem for bag-of-words approaches. Thus, they used a Convolutional Neural Network, where each word is represented by its embedding and each named entity is replaced by its tag, e.g., \emph{person}, \emph{organization}, \emph{location}. 

Unlike the above work, we mimic the selection strategy of \emph{one}
specific fact-checking organization by learning to jointly predict the selection choices by \emph{multiple} such organizations.

\section{Data}
\label{sub:data}

In our experiments, we used the CW-USPD-2016 dataset from our previous work \cite{gencheva-EtAl:2017:RANLP}, which can be found on GitHub.\footnote{\url{http://github.com/pgencheva/claim-rank}}
It is derived from transcripts of the 2016 US Presidential campaign, and includes one Vice-Presidential and three Presidential debates, all of which were fact-checked by the following nine reputable fact-checking organizations: PolitiFact, FactCheck, ABC, CNN, NPR, NYT, Chicago Tribune, The Guardian, and The Washington Post.

Overall, there are four debates with a total of 5,415 sentences. A sentence is considered check-worthy with respect to a source if that source has chosen to fact-check it. Overall, a total of 880 sentences were fact-checked by at least one source, 191 were selected by two or more sources, 100 by three or more, and only one sentence was chosen by all nine sources, as Table~\ref{table:agreement} shows. Table~\ref{table:exampleA} shows an example: interventions by Hillary Clinton and Donald Trump from the first US presidential debate. This reflects the disparities in check-worthiness selection criteria. More details about the dataset can be found in~\cite{gencheva-EtAl:2017:RANLP}.

\begin{table}[h]
\centering
\small
  \begin{tabular}{crr}
\toprule
    \bf Selected by & \bf Number of & \bf Cumulative \\
    \bf \# Sources & \bf Sentences & \bf Sum \\
    \midrule
    9 & 1 & 1 \\
    8 & 6 & 7 \\
    7 & 5 & 12 \\
    6 & 19 & 31 \\
    5 & 26 & 57 \\
    4 & 40 & 97 \\
    3 & 100 & 197 \\
    2 & 191 & 388 \\
    \bf 1 & \bf 492 & \bf 880 \\
    \bottomrule
  \end{tabular}
\caption{Agreement between the fact-checkers: sentences selected by 1, 2, $\ldots$, 9 of them.}
\label{table:agreement}
\end{table}

\begin{figure}[h]
\includegraphics[width=225pt]{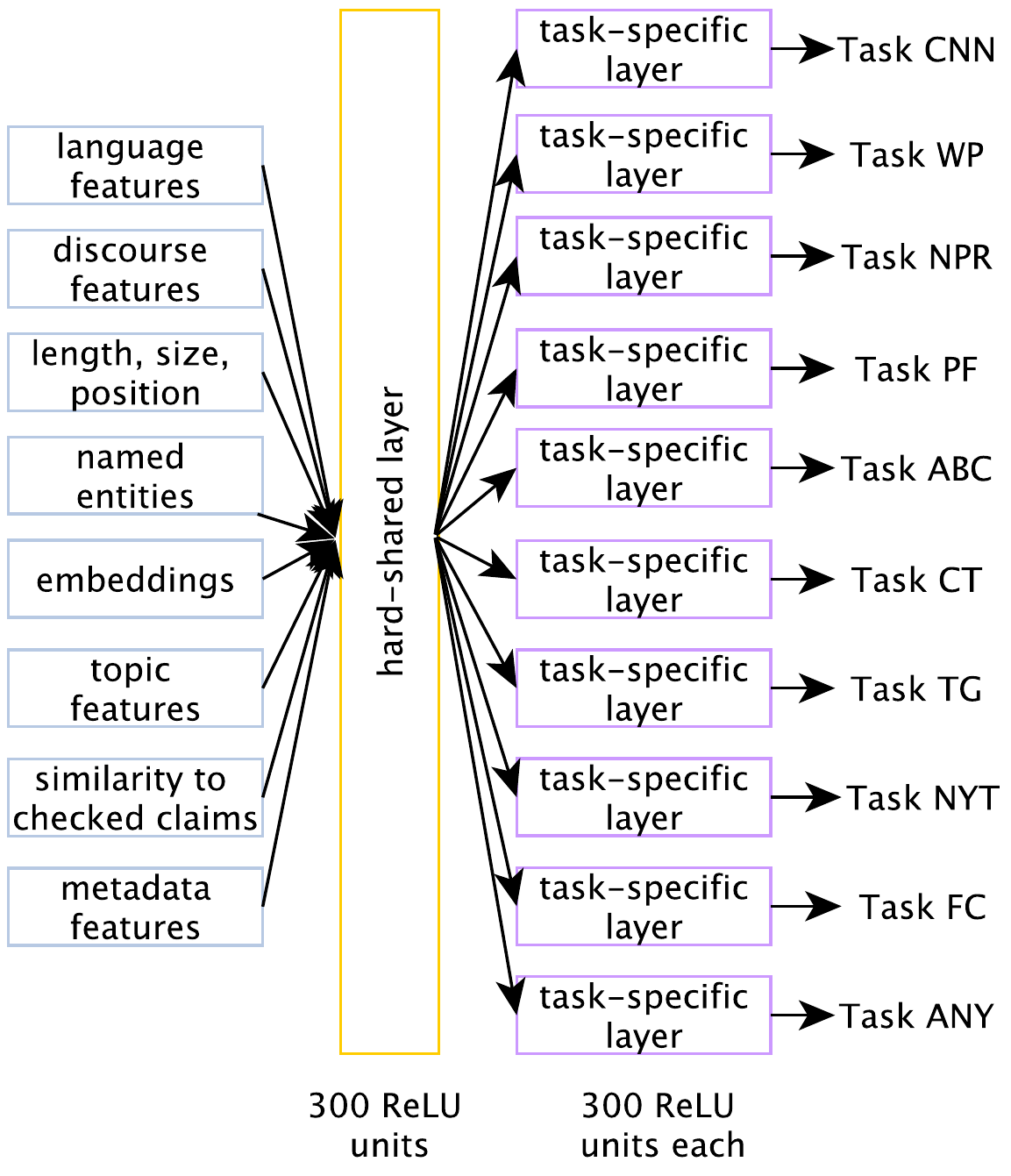}
\caption{The architecture of our neural multi-task learning model, predicting whether each of the nine individual fact-checking organizations (tasks) would consider this sentence check-worthy and one cumulative source: \emph{task ANY}.}
\label{figure:architecture}
\end{figure}

\section{Our Multi-Task Learning Model}
\label{sec:method}

We approach the task of check-worthiness prediction as a multi-source learning problem, using different sources of annotation over the same training dataset. Thus, we can learn to mimic the selection strategy of each of the individual sources.

Figure~\ref{figure:architecture} shows the architecture of our neural multi-task learning model which, given an input sentence in the context of a political debate, predicts whether each of the nine individual sources (tasks) would have selected it, and whether at least one of them would, which is the special \emph{task ANY}.

The input to our neural network consists of various domain-specific features that have been previously shown to work well for the task of check-worthiness prediction.
In particular, from~\cite{Hassan:15}, we adopt TF.IDF-weighted bag of words, 
part-of-speech tags,
the presence of named entities,
sentiment scores,
and sentence length (in number of tokens).
Moreover, from~\cite{gencheva-EtAl:2017:RANLP}, we further adopt \emph{lexicon features}, e.g.,~for bias~\cite{Recasens+al:13a}, for sentiment~\cite{Liu:2005:OOA:1060745.1060797}, for assertiveness~\cite{hooper1974assertive}, and for subjectivity; \emph{structural features}, e.g., for location of the sentence within the debate/intervention; LDA topics~\cite{blei2003latent}; word embeddings, pretrained on Google News~\cite{mikolov-yih-zweig:2013:NAACL-HLT}; and discourse relations with respect to the neighboring sentences~\cite{jotycodra}. See~\cite{Hassan:15,gencheva-EtAl:2017:RANLP} for more details about each of these feature types.  

After the input layer, comes a hidden layer that is shared between all tasks. It is followed by ten parallel task-specific hidden layers. During training, in the process of backpropagation, each task modifies the weights of its own task-specific layer and also of the shared layer. 

Finally, each task-specific layer is followed by an output layer: a single sigmoid unit that provides the prediction of whether the utterance was fact-checked by the corresponding source. Eventually, we make use of the probability of the prediction to prioritize claims for fact-checking.
This kind of neural network architecture for multi-task learning is known in the literature as \emph{hard parameter sharing} \cite{Caruana93multitasklearning}, and it can greatly reduce the risk of overfitting.

\begin{table*}[t]
\footnotesize
\small
\centering
\begin{tabular}{l@{\hspace{1mm}} c@{\hspace{1mm}} c@{\hspace{1mm}} c@{\hspace{1mm}} c@{\hspace{1mm}} c@{\hspace{1mm}} c@{\hspace{1mm}} c@{\hspace{1mm}} c@{\hspace{1mm}} c@{\hspace{1mm}} c@{\hspace{2mm}} p{7.5cm}}
\hline
\bf Speaker & \bf Total & \bf CT& \bf ABC & \bf CNN & \bf WP& \bf NPR& \bf PF& \bf TG & \bf NYT & \bf FC& \bf Text\\ 
\hline
Clinton &   0 & 0 & 0 & 0 & 0 & 0 & 0 & 0 & 0 & 0 & So we're now on the precipice of having a potentially much better economy, but the last thing we need to do is to go back to the policies that failed us in the first place. \\
\blue{Clinton} &    \blue 6 &   \blue 1 &   \blue 1 &   \blue 0 &   \blue 0 &   \blue 1 &   \blue 1 &   \blue 0 &   \blue 1 &   \blue 1 &   \blue{Independent experts have looked at what I've proposed and looked at what Donald's proposed, and basically they've said this, that if his tax plan, which would blow up the debt by over \$5 trillion and would in some instances disadvantage middle-class families compared to the wealthy, were to go into effect, we would lose 3.5 million jobs and maybe have another recession.} \\
\blue{Clinton} &    \blue 1 &   \blue 1 &   \blue 0 &   \blue 0 &   \blue 0 &   \blue 0 &   \blue 0 &   \blue 0 &   \blue 0 &   \blue 0 &   \blue{They've looked at my plans and they've said, OK, if we can do this, and I intend to get it done, we will have 10 million more new jobs, because we will be making investments where we can grow the economy.}\\
Clinton &   0 & 0 & 0 & 0 & 0 & 0 & 0 & 0 & 0 & 0 & Take clean energy.\\
Clinton &   0 & 0 & 0 & 0 & 0 & 0 & 0 & 0 & 0 & 0 & Some country is going to be the clean- energy superpower of the 21st century.\\
\blue{Clinton} &    \blue 6 &   \blue 1 &   \blue 1 &   \blue 1 &   \blue 1 &   \blue 0 &   \blue 0 &   \blue 1 &   \blue 0 &   \blue 1 &   \blue{Donald thinks that climate change is a hoax perpetrated by the Chinese.}\\
Clinton &   0 & 0 & 0 & 0 & 0 & 0 & 0 & 0 & 0 & 0 & I think it's real.\\
\blue{Trump} &  \blue 5 & \blue 1 & \blue 1 &   \blue 0 &\blue 1 &  \blue 1 &   \blue 1 &   \blue 0 &   \blue 0 &   \blue 0 &   \blue{I did not.}\\
\hline
 \end{tabular}
\caption{Excerpt from the transcript of the first US 2016 Presidential Debate, annotated by nine sources: Chicago Tribune, ABC News, CNN, Washington Post, NPR, PolitiFact, The Guardian, The New York Times and Factcheck.org. Whether the media fact-checked the claim or not is indicated by a 1 or 0, respectively. The blue sentences are considered as positive in the \textit{any} setting.}
\label{table:exampleA}
\end{table*}

\section{Experiments and Evaluation}
\label{sec:experiments}

As the CW-USPD-2016 corpus contains four debates, we perform 4-fold cross-validation, where each time we leave one debate out for testing, and we train on the remaining three debates. Moreover, in order to stabilize the results, we repeat each experiment three times with different random seeds and we report the average over these three reruns of the system.\footnote{Having multiple reruns is a standard procedure to stabilize an optimization algorithm that is sensitive to the random seed, e.g., this strategy has been argued for when using MERT for tuning hyper-parameters in Statistical Machine Translation~\cite{Foster:2009:SME}.}

\noindent In our neural model, we used ReLU units and a shared layer of size 300. For training, we used Stochastic Gradient Descent with Nesterov momentum,\footnote{Using Adam optimizer was faster, converging after only 30 epochs, but it yielded slightly worse results.}
iterating for 100 epochs.

\renewcommand{\tabcolsep}{2pt}
\begin{table*}
\small
\begin{minipage}[t]{0.5\textwidth}
\vspace{0pt}
\begin{tabular}[t]{l@{}rrrrrr}
\toprule
    \bf Model & \bf MAP & \bf R-Pr & \bf P@5 & \bf P@10 & \bf P@20 & \bf P@50 \\
    \hline
\\
\multicolumn{7}{l}{ \bf ABC} \\
\hline
    \mCBfeats & .057 & .061 & .050 & .038 & .056 & .050\\
    \mCBP & .065 &.066& .150 & .125 & .088 & .080\\ 
    \mRANLPsingle & .059 & .068 & .050 & .050 & .100 & .060 \\ 
    \hline
      \mA & \underline{.097} & \underline{.112} & \underline{.250} & \underline{.175} & \underline{.162} & \underline{.100} \\
  \bf \mD & \underline{\bf .119} & \underline{\bf .157} & \underline{\bf .333} & \underline{\bf .225} & \underline{\bf .217} & \underline{\bf .122} \\
  \bf \mE & \underline{\bf .118} & \underline{\bf .160} & \underline{\bf .300} & \underline{\bf .233} & \underline{\bf .229} & \underline{\bf .132} \\
\midrule
\\
\multicolumn{7}{l}{ \bf The Washington Post (WP)} \\
\hline
    \mCBfeats & .051 & .053 & .050 & .033 & .046 & .048\\
    \mCBP & .048 &.056& .050 & .075 & .050 & .045\\ 
    \mRANLPsingle & .102&.098 & .200 & .175 & .113 & .080 \\ 
    \hline
      \mA & \underline{.106} & \underline{.110} & .150 & .100 & .112 & \underline{.110} \\
  \bf \mD & \underline{\bf .127} & \underline{\bf .127} & \underline{\bf .350} & \underline{\bf .233} & \underline{\bf .162} & \underline{\bf .123} \\
  \bf \mE & \underline{\bf .130} & \underline{\bf .129} & \underline{\bf .350} & \underline{\bf .250} & \underline{\bf .171} & \underline{\bf \emph{.110}} \\
\midrule 
\\
\multicolumn{7}{l}{ \bf CNN} \\
\hline
    \mCBfeats & .055 & .058 & .063 & .038 & .050 & .053\\
    \mCBP & .082 & .096& .150 & .125 & .088 & .085\\ 
    \mRANLPsingle & .079 & .076 & .100 & .100 & .100 & .090 \\ 
\hline
\mA & \underline{.087} & \underline{.091} & \underline{.250} & \underline{.150} & \underline{.121} & .090 \\
  \bf \mD & \bf \underline{.113} & \bf \underline{.132} & \bf \emph{\underline{.250}} & \bf \underline{.208} & \bf \underline{.183} & \bf \underline{.140} \\
  \bf \mE & \bf \underline{.109} & \bf \underline{.126} & \underline{.167} & \bf \underline{.200} & \bf \underline{.167} & \bf \underline{.128} \\
\midrule
\\
\multicolumn{7}{l}{ \bf FactCheck (FC)} \\
\hline
    \mCBfeats & .068 & .072 & .108 & .071 & .077 & .070\\
    \mCBP & .081&.213 & .150 & .125 & .100 & .115\\
    \mRANLPsingle & .081&.098 & .050 & .125 & .088 & .085 \\ 
\hline
      \mA & \underline{.084} & \underline{.114} & \underline{.117} & .125 & .088 & \underline{.100} \\
  \bf \mD & \bf \underline{.105} & \bf \underline{.136} & \bf \underline{.250} & \bf \underline{.175} & \bf \underline{.146} & \bf \underline{.118} \\
  \bf \mE & \bf \underline{.117} & \underline{.110} & \bf \underline{.333} & \bf \underline{.242} & \bf \underline{.196} & \bf \underline{.107} \\
\midrule
\\
\multicolumn{7}{l}{\bf PolitiFact} \\
\hline
    \mCBfeats & .137 & .143 & .250 & .200 & .188 & .185\\
    \mCBP & .154 &.213 & .200 & .300 & .238 & .210\\ 
    \mRANLPsingle & .218 & .274 & .450 & .325 & .300 & .270 \\ 
    \hline
           \mA & .201 & \underline{.278} & .250 & .250 & .262 & .262 \\
  \textbf{\mD} & \bf .209 & .258 & \bf .400 & \bf \underline{.367} & \bf \underline{.317} & \bf .270 \\
  \textbf{\mE} & \bf .210 & .252 & \bf \underline{.500} & \bf \underline{.350} & \bf \underline{.333} & \bf \underline{.272} \\ 
\bottomrule 
\end{tabular}
\end{minipage} \hfill
\begin{minipage}[t]{0.5\textwidth}
\strut\vspace*{-\baselineskip}\newline
\begin{tabular}[t]{l@{}rrrrrr}
\toprule
    \bf Model & \bf MAP & \bf R-Pr & \bf P@5 & \bf P@10 & \bf P@20 & \bf P@50 \\
    \hline
    \\
\multicolumn{7}{l}{ \bf NPR} \\
\hline
    \mCBfeats & .079 & .085 & .136 & .089 & .096 & .087\\
    \mCBP & .144 & .186 &.200 & .225 & .225 & .180\\ 
    \mRANLPsingle & .193 & .216& .550 & .475 & .350 & .255 \\ 
\hline
      \mA & .175 & .195 & .250 & .250 & .283 & .228 \\
  \bf \mD & \bf .186 & \bf .210 & \bf .333 & \bf .342 & \bf .300 & \bf .245 \\
  \bf \mE & \bf .180 & \bf .207 & \bf .333 & \bf .283 & .250 & .227 \\
\midrule
\\
\multicolumn{7}{l}{ \bf The Guardian (TG)} \\
\hline
    \mCBfeats & .066 & .075 & .110 & .070 & .070 & .066\\
    \mCBP & .084 &.128& .100 & .100 & .125 & .140\\
    \mRANLPsingle & .121&.156 & .250 & .225 & .200 & .155 \\ 
\hline
      \mA & \underline{.127} & \underline{.174} & .200 & .150 & .196 & \underline{.178} \\
  \bf \mD & \bf \underline{.133} & \bf \underline{.199} & .183 & \bf .175 & \bf .192 & \bf \underline{.193} \\
  \bf \mE & \bf \underline{.130} & \underline{.159} & \bf \bf .217 & \bf .175 & \bf .200 & \underline{.167} \\
\midrule
\\
\multicolumn{7}{l}{ \bf Chicago Tribune (CT)} \\
\hline
    \mCBfeats & .058 & .063 &.050 & .050 & .050 & .065\\
    \mCBP & .053 &.032& .050 & .050 & .038 & .065\\ 
   \mRANLPsingle & .087&.118 & .150 & .150 & .175 & .105 \\ 
\hline
      \mA & .079 & .110 & .100 & .100 & .125 & .075 \\
  \bf \mD & \bf .081 & .090 & \bf \emph{.100} & \bf .133 & .104 & \bf .082 \\
  \bf \mE & \bf .087 & .087 & \bf .133 & \bf \emph{.100} & .108 & \bf .093 \\
\midrule
\\
\multicolumn{7}{l}{ \bf The New York Times (NYT)} \\
\hline
    \mCBfeats & .080 & .084 & .138 & .094 & .100 & .088\\
    \mCBP & .103 & .250 & .250 & .163 & .135 & .135\\ 
    \mRANLPsingle & .136 & .178 & .250 & .225 & .188 & .135 \\
\hline
      \mA & \underline{.187} & \underline{.221} & \underline{.350} & \underline{.325} & \underline{.238} & \underline{.192} \\
  \bf \mD & \underline{.150} & \underline{.213} & .233 & .200 & \underline{.196} & \underline{.180} \\
  \bf \mE & \underline{.147} & \underline{.197} & .200 & .167 & .158 & \underline{.162} \\
\midrule
\\

\\
\hline
    \mCBfeats &\multicolumn{6}{l}{Singleton only w/ClaimBuster features}\\
    \mCBP &\multicolumn{6}{l}{Online version of ClaimBuster}\\
    \mRANLPsingle  &\multicolumn{6}{l}{Singleton from~\cite{gencheva-EtAl:2017:RANLP}}\\
\hline
      \mA   & \multicolumn{6}{l}{Trained on the target medium only} \\
  \bf \mD & \multicolumn{6}{l}{Multi-task for nine sources} \\
  \bf \mE & \multicolumn{6}{l}{Multi-task for nine sources+any} \\
\bottomrule

  \end{tabular}
  \end{minipage} \hfill
\caption{Evaluation results for each of the nine fact-checking sources as a target to mimic. Shown are the results for single-source baselines vs. for multi-task learning with nine and with ten classes. The improvements over the singleton baseline are marked in bold. We further compare to \emph{singleton} that is limited to ClaimBuster's features, to the online version of ClaimBuster \cite{Hassan:15}, and to \emph{\mRANLPsingle} results in \cite{gencheva-EtAl:2017:RANLP}. The improvements over the latter are underlined.}
\label{table:multitask:results}
\end{table*}

Recall that our main objective is to prioritize the claims that should be selected for manual fact-checking, which is best achieved by proposing a ranked list of claims.
Thus, we have a ranking task, for which we use suitable information retrieval evaluation measures. In particular, we adopt Mean Average Precision (MAP) as our primary evaluation measure.
We further report R-Precision, or R-Pr, and precision at $k$, or P@$k$,\footnote{See \cite{Buckley:2000} for a discussion on these evaluation measures.} for $k=\{5, 10, 20, 50\}$. Note that 50 is the approximate number of claims checked by most of the sources for each debate (the exception being PolitiFact, with up to 99 checked claims).

Table~\ref{table:multitask:results} presents the evaluation results comparing three models. The first one is a single-task model \mA where a separate neural network is trained for each source. 
The other two are multi-task learning models:
\mD predicts labels for each of the nine tasks, one for each fact-checker,
and \mE predicts labels for each of the nine tasks (one for each fact-checker), and also for \emph{task ANY} (as shown in Figure~\ref{figure:architecture}).
We further compare to the online version of ClaimBuster \cite{Hassan:15} and to the \emph{singleton} results reported in~\cite{gencheva-EtAl:2017:RANLP} (on the same dataset, with the same cross-validation).\footnote{Note that we could not compare to \cite{Patwari:17} directly as they used a different dataset. However, they use a small set of basic features that overlap with those of ClaimBuster \cite{Hassan:15} to a large extent, and thus we expect that they would perform similarly to ClaimBuster.}

We can see in Table~\ref{table:multitask:results} that our \mA is comparable and even slightly better than the \mA model in \cite{gencheva-EtAl:2017:RANLP}, and both outperform the online version of ClaimBuster \cite{Hassan_thequest:2015}.
We further see that limiting our singleton system to ClaimBuster's features yields a sizable drop in performance.
Moreover, for most sources, multi-task learning improves over the singleton models. 
The results of the multi-task variations that improve over the single baseline are boldfaced.
The improvements are consistent across the evaluation measures, but they vary largely depending on the fact-checking source and the evaluation measure.

\noindent One notable exception is NYT, for which the single-task learning shows the highest scores. We hypothesize that the network has found some distinctive features of NYT, which make it easy to predict. These relations are blurred when we try to optimize for multiple tasks at once. However, it is important to state that removing NYT from the learning targets worsens the results for the other sources, i.e.~it carries some important relations that are worth modeling.

Table~\ref{tab:avg-map} presents the same results but averaged over the nine sources. 
The first section in Table~\ref{tab:avg-map} shows the results for the online version of ClaimBuster \cite{Hassan:15}, and for the \emph{singleton} and the \emph{task ANY} results in \cite{gencheva-EtAl:2017:RANLP}. We can see that our \mA model is comparable to the \mA and \mB models in \cite{gencheva-EtAl:2017:RANLP}, and our multi-task learning models consistently improve over them for all evaluation measures in all but one case.

It is common in neural networks to try to implicitly learn the representations based on word embeddings. We include this as a baseline in the second section in Table~\ref{tab:avg-map}. The performance of the model that only uses embeddings is in general poor, which suggests that complex feature modeling is necessary for this task; including features that go beyond the current-sentence level. Further feature analysis is included in Table~\ref{tab:feature-importance}.

\noindent The third section of Table~\ref{tab:avg-map} presents the results for the models of this paper. Again, we can see that multi-task learning yields sizeable improvement over the single-task learning baseline for all evaluation measures. 

Another conclusion that can be drawn from this table is that including the task \emph{task ANY} (i.e.,~whether any of the nine media would select a target) does not help to improve the multi-task model. This is probably due to the fact that this information is already contained in the multi-task model with nine sources.

The last section in Table~\ref{tab:avg-map} presents two additional variants of the model: the single-task learning \mB system ---which trains on the union of the selected sentences by all nine fact-checkers to predict the target fact-checker only---, and 
the system \mC that predicts labels for two tasks: (\emph{i})~for the target fact-checker, and (\emph{ii})~for \emph{task ANY}. We can see that \mB performs comparably to the \mA baseline, thus being clearly inferior than the multi-task learning variants. 
Finally, \mC is also better than the single-task learning variants, but it falls short compared to the other multi-task learning variants. Including output units for all nine individual media seems crucial for getting advantage of the multi-task learning, i.e.,~considering only an extra output prediction node for the \emph{task ANY} problem is not enough. 

\renewcommand{\tabcolsep}{2pt}
\begin{table}[tbh]
\centering
\small
\begin{tabular}{l@{}rrrrrr}
\toprule
\bf Model & \bf MAP & \bf R-Pr & \bf P@5 & \bf P@10 & \bf P@20 & \bf P@50 \\
\midrule
\mCBP & .090 & .138 & .144 & .143 & .121 & .117\\
\mRANLPsingle & .120 & .142 & .228 & .206 & .179 & .137\\
\mRANLP & .128 & .225 & .194 & .186 & .178 & .153\\
\hline
\mA \emph{(embed.)} 
& .058 & .065 & .055     & .055    & .068 & .072\\
    \mCBfeats & .072 & .077 & .106 & .076 & .081 & .079\\
\hline
      \mA & \underline{.127} & \underline{.156} & .213 & .181 & .176 & \underline{.148} \\
  \bf \mD & \bf \underline{.136} & \bf \underline{.169} & \bf \underline{.270} & \bf \underline{.229} & \bf \underline{.202} & \bf \underline{.164} \\
  \bf \mE & \bf \underline{.136} & \bf \underline{.159} & \bf \underline{.281} & \bf \underline{.222} & \bf \underline{.201} & \bf \underline{.155} \\
  \hline
      \mB & \underline{.125} & \underline{.153} & .204 & .197 & .175 & \underline{.153} \\    
      \mC & \bf \underline{.130} & \underline{.153} & \bf \underline{.237} & \bf \underline{.220} & \bf \underline{.184} & \bf \emph{\underline{.148}} \\
\bottomrule
\end{tabular}

\caption{Evaluation results averaged over nine fact-checking organizations (see Table~\ref{table:multitask:results} for the unrolled results). We compare multi-task learning to three \emph{singleton} baselines; the improvements are shown in bold. The first section compares to the online version of ClaimBuster \cite{Hassan:15}, as well as to \emph{singleton} and to \emph{task ANY} results in \cite{gencheva-EtAl:2017:RANLP}. The improvements over the latter are underlined. The last section shows the results for two more baselines: \mB and \mC.}
\label{tab:avg-map}
\end{table}

\section{Discussion}
\label{sec:discuss}

In this section, we provide deeper insight into the peculiar characteristics of the multi-task model. 
\medskip

\noindent
\textbf{Error Analysis} First, we perform comparative error analysis, showing both examples of improvement of the proposed \mD model with respect to the \mA as well as some cases where the former fails.
The results are shown in Table~\ref{tab:err-analysis}. The first four rows are true positive claims, which were misclassified by the \emph{singleton} model, but were correctly classified by the \emph{multi-task} one. As we can see, the claims were selected for fact-checking by many organizations: between six and eight. This reflects that these instances were certainly check-worthy and the multi-task model correctly spotted them.
The observation holds for a prevailing number of all of the new true positives. This is a natural consequence of our neural architecture, where all sources share a hidden layer and tend to learn from the selection criteria of the other sources as well.

\begin{table}[t]
\small
\begin{tabular}{r@{\hspace{1mm}}cc@{\hspace{1mm}}cp{57mm}}
\toprule
\bf N & \bf Type & \bf Tgt & \bf \# & \bf Sentence \\   \hline
1 & TP & CT & 8 & \textbf{Trump} $\blacktriangleright$ It's gone, \$6 billion.\\
2 & TP & WP & 8 & \textbf{Trump} $\blacktriangleright$ I was against -- I was against the war in Iraq. \\
3 & TP & TG & 6 & \textbf{Trump} $\blacktriangleright$ You ran the State Department, \$6 billion was either stolen. \\
4 & TP & NYT & 6 & \textbf{Pence} $\blacktriangleright$ Less than 10 cents on the dollar of the Clinton Foundation has gone to charitable causes. \\\hline
5 & FP & CT & 4 & \textbf{Trump} $\blacktriangleright$ Wrong. \\
6 & FP & CT & 3 & \textbf{Trump} $\blacktriangleright$ In Chicago, they've had thousands of shootings, thousands since January 1st. \\
7 & FP & CNN & 0 & \textbf{Clinton} $\blacktriangleright$ Donald has said he's in favor of defending Planned Parenthood. \\
8 & FP & WP & 0 & \textbf{Trump} $\blacktriangleright$ I never met Putin. \\    \hline
9 & FN & FC & 6 & \textbf{Clinton} $\blacktriangleright$ Donald thinks that climate change is a hoax perpetrated by the Chinese. \\
10 & FN & NYT & 4 & \textbf{Pence} $\blacktriangleright$ And Iraq has been overrun by ISIS, because Hillary Clinton failed to renegotiate... \\
11 & FN & NPR & 1 & \textbf{Trump} $\blacktriangleright$ China should go into North Korea. \\
12 & FN & NPR & 1 & \textbf{Trump} $\blacktriangleright$ We have no growth in this country. \\
\bottomrule
\\
\end{tabular}

\caption{Sentences with prediction type (for the \mD model, with respect to the target medium), the target medium, and total number of media that selected this sentence (\#).}
\label{tab:err-analysis}
\end{table}

Two types of false positive errors occur in rows 5--8. Rows 5 and 6 are predicted by multiple sources that reinforce one another for the wrong guess. We can attribute this to the specifics of the multi-task architecture. On the one hand, the shared layer helps a medium to learn from the selection process of other media. On the other hand, it begins to make more mistakes on claims selected by more media.

\noindent On the contrary, rows 7 and 8 show claims that are not check-worthy for any source, but exhibit features such as named entities and negations that typically suggest that the claim might be check-worthy.
Finally, rows 9--12 are false negative instances. We have two claims that were fact-checked by several media and two selected by one medium only. The first group indicates that some tasks might try to learn their own features, while the second group shows a possible down side of the multi-task model.
\medskip 

\noindent
\textbf{Feature Importance} 
Next, we conduct feature ablation experiments to determine which of the feature groups are most important for the final multi-task model. For this purpose, we remove one feature group at a time from the \mD model. 

\begin{table}[tbhp]
\centering
\small
\begin{tabular}{l@{}cccccc}
\toprule
\bf Feature & \bf MAP & \bf R-Pr & \bf P@5 & \bf P@10 & \bf P@20 & \bf P@50 \\
\midrule
Embeddings & .102 & .133 & .250 & .231 & .188 & .129 \\
Metadata & .120 & .147 & .278 & .217 & .175 & .139 \\
Sentiment & .122 & .146 & .233 & .203 & .164 & .140 \\
Topics & .123 & .147 & .244 & .211 & .172 & .142 \\
Discourse & .123 & .140 & .261 & .217 & .175 & .141 \\
NER & .125 & .149 & .244 & .217 & .178 & .140 \\
Segment size & .125 & .149 & .256 & .211 & .172 & .139 \\
Position & .125 & .143 & .261 & .219 & .193 & .138 \\
Linguistic & .126 & .150 & .250 & .208 & .190 & .151 \\
Contradiction & .126 & .149 & .250 & .203 & .174 & .142 \\
Lengths & .127 & .144 & .272 & .233 & .175 & .147 \\
Sim. to prev. & .127 & .151 & .222 & .214 & .178 & .148 \\
\bottomrule
\end{tabular}
\caption{Ablation experiments: removing a feature group from the \mD model, using all nine tasks.}
\label{tab:feature-importance}
\end{table}

Table~\ref{tab:feature-importance} shows that without the Embedding features the performance of the model drops significantly. They were also the best features in the \emph{\mRANLPsingle} model of~\newcite{gencheva-EtAl:2017:RANLP}. Metadata features are the second most important for the model. An interesting observation is that some of the best-preforming features from \emph{\mRANLPsingle} are the least contributing to the multi-task model. Such features are \emph{Sim. to prev.} (similarity to previously fact-checked claims), and the linguistic features.

\medskip

\begin{figure}[t]
\centering
\includegraphics[width=240pt]{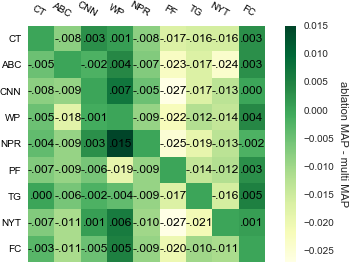}
\caption{Ablation experiment with the \mD model. Each row is an experiment removing one target. Each column is the MAP \textit{difference} with respect to the \mD model for the corresponding target.}
\label{figure:ablation}
\end{figure}

\noindent
\textbf{Source Ablation}
Figure~\ref{figure:ablation} shows ablation results with the \mD model. A cell at row $r$ and column $c$ shows the performance difference for target $c$ when excluding the target $r$ at training time. For example, in the first row we run the \mD model neglecting CT in the set of targets. 
Negative values indicate that removing target $r$ worsens the MAP of target $c$. Conversely, positive values indicate that removing target $r$ improves MAP for target $c$. 
We can observe that the MAP of ABC has dropped by .008, meaning that ABC finds beneficial information from sharing a layer with the CT target. On the contrary, the target FC improves after removing CT, pointing out the presence of conflicts in the learning phase of the shared layer. The largest decrease in MAP is observed in PF after removing CNN, NYT, and NPR. On the other hand, the most significant increase in MAP is in WP after removing NPR and CNN.

\section{Conclusion and Future Work}
\label{sec:conclusion}

We have presented a multi-task learning approach for estimating the check-worthiness of claims in political debates, and we have further demonstrated its effectiveness experimentally, pushing the state of the art. 

In future work, we plan to experiment with more debates. We further plan to go beyond debates, i.e., to general news articles. Moreover, we would like to apply our approach to other languages for which multiple check-worthiness annotations of the same dataset are available. 

We plan to try information sources such as the Web \cite{Popat:2017:TLE:3041021.3055133}, as well as tweets and temporal information \cite{ma2016detecting}.
We also want to explore other multi-task learning options, e.g., as described in \cite{DBLP:journals/corr/Ruder17a}.

It would be interesting to investigate the reasons why the NYT source does not benefit from the multi-task architecture. In order to adapt to this situation with a single model, 
we plan to experiment with a network with \emph{soft parameter sharing}, e.g.,~as in~\cite{duong2015}.
For example, we could create a chain of layers that back-propagate to the input using only single task targets and then add an auxiliary layer that is shared between the tasks on the side. In this way, the model would be able to turn off the multi-task learning completely for some of the sources. However, training such kind of model might require significantly more training data; semi-supervised training might be a possible solution.

\section*{Acknowledgments}

We would like to thank the anonymous reviewer, whose constructive feedback has helped us improve the quality of this paper.

\medskip

This work is part of the Tanbih project,\footnote{\url{http://tanbih.qcri.org/}} which aims to limit the effect of ``fake news'', propaganda and media bias by making users aware of what they are reading. The project is developed in collaboration between the Qatar Computing Research Institute (QCRI), HBKU and the MIT Computer Science and Artificial Intelligence Laboratory (CSAIL).

\bibliography{acl2019}
\bibliographystyle{acl_natbib}

\appendix

\end{document}